\documentclass[conference]{IEEEtran}
\IEEEoverridecommandlockouts
\usepackage{cite}
\usepackage{amsmath,amssymb,amsfonts}
\usepackage{algorithmic}
\usepackage{graphicx}
\usepackage{textcomp}
\usepackage{xcolor}
\usepackage{relsize}

\def\BibTeX{{\rm B\kern-.05em{\sc i\kern-.025em b}\kern-.08em
    T\kern-.1667em\lower.7ex\hbox{E}\kern-.125emX}}
    
\usepackage{amsthm}

\allowdisplaybreaks

\begin{document}

\title{\huge{Bayesian Simultaneous Localization and Multi-Lane Tracking Using Onboard Sensors and a SD Map}\\
\thanks{This work was partially supported by the Wallenberg AI, Autonomous Systems and Software Program (WASP) funded by the Knut and Alice Wallenberg Foundation.}
}
\author{Yuxuan Xia$^\dagger$$^\star$, Erik Stenborg$^\dagger$, Junsheng Fu$^\dagger$, and Gustaf Hendeby$^\star$\\
\small{$^\dagger$Zenseact, Gothenburg, Sweden}\\
\small{$^\star$Department of Electrical Engineering, Link\"{o}ping University, Link\"{o}ping, Sweden}\\
\small{Email: $^\dagger$firstname.lastname@zenseact.com, $^\star$firstname.lastname@liu.se}\vspace{-5mm}
}
\maketitle

\begin{abstract}

% High-definition maps with accurate lane-level information is crucial for autonomous driving. However, the creation of such maps is an expensive and resource-intensive process. To this end, we present a cost-effective solution to create lane-level maps using only a global navigation satellite system (GNSS) and a camera. Leveraging a prior standard-definition (SD) map, our proposed solution utilizes GNSS measurements and vision data, including visual odometry, traffic sign bounding boxes, and lane marking detection points, to simultaneously estimate the vehicle's 6D pose, its position within a SD map, poses and sizes of traffic signs, as well as the geometries of traffic lines. This is achieved using a Bayesian simultaneous localization and multiple object tracking filter, where the estimation of traffic lines conditioned on the vehicle's pose is formulated as a multiple extended object tracking problem and addressed using a trajectory Poisson multi-Bernoulli mixture (TPMBM) filter. In TPMBM filtering, traffic lines are modeled using B-spline trajectories, each parameterized by a sequence of control points, and a Bernoulli object spawning model is used to handle lane splitting. The proposed solution has been evaluated using real-world data, and promising results are obtained.

High-definition map with accurate lane-level information is crucial for autonomous driving, but the creation of these maps is a resource-intensive process. To this end, we present a cost-effective solution to create lane-level roadmaps using only the global navigation satellite system (GNSS) and a camera on customer vehicles. Our proposed solution utilizes a prior standard-definition (SD) map, GNSS measurements, visual odometry, and lane marking edge detection points, to simultaneously estimate the vehicle's 6D pose, its position within a SD map, and also the 3D geometry of traffic lines. This is achieved using a Bayesian simultaneous localization and multi-object tracking filter, where the estimation of traffic lines is formulated as a multiple extended object tracking problem, solved using a trajectory Poisson multi-Bernoulli mixture (TPMBM) filter. In TPMBM filtering, traffic lines are modeled using B-spline trajectories, and each trajectory is parameterized by a sequence of control points. The proposed solution has been evaluated using experimental data collected by a test vehicle driving on highway. Preliminary results show that the traffic line estimates, overlaid on the satellite image, generally align with the lane markings up to some lateral offsets.

\end{abstract}

\vspace{-5mm}

\section{Introduction}
High-definition (HD) maps with detailed lane-level information are an enabling factor for autonomous driving, providing the necessary precision, safety, and efficiency in localization, navigation, and decision-making processes \cite{elghazaly2023high}. However, the creation of HD maps is an expensive and resource-intensive process, which typically involves using vehicles equipped with high-precision equipment, such as lidar, to continuously collect and process the data. An appealing, cost-effective alternative is to instead use onboard sensors available in massive customer vehicles, such as the global navigation satellite system (GNSS) and camera, to jointly create large-scale, lane-level geometric maps. This approach is known as crowdsourced mapping \cite{zhou2022lane}, whereby multiple sources of low-fidelity data are aggregated to create high-fidelity maps.

In this paper, we focus on the problem of creating lane-level geometric maps, which are built upon standard-definition (SD) maps with low geometric fidelity, using single-antenna GNSS measurements and mono camera data collected from a single vehicle traversal. This can serve as a solid foundation for the follow-up map change detection, update, and merging tasks. In particular, we aim to estimate the 3D geometry of traffic lines using lane marking edge detection points obtained from a computer vision algorithm. Since the accurate estimation of traffic line geometries relies on the precise vehicle pose, which could hardly be obtained using only GNSS, the vehicle's pose and the traffic line geometries need to be simultaneously estimated. To this end, we propose an effective solution that leverages a prior SD map, GNSS measurements, visual odometry, and lane marking edge detection points to simultaneously estimate the vehicle's 6D pose, its position within a SD map, and the 3D geometry of traffic lines.

Our proposed solution is achieved using a Bayesian simultaneous localization and multi-object tracking (SLAMOT) filter, where the estimation of traffic lines is formulated as a multiple extended object tracking (EOT) problem. Note that, although traffic lines remain in fact static, in most cases they can only be partially observed by the camera due to its limited field-of-view. Thus, a dynamic model for the traffic lines is needed to make predictions of their unseen geometries before each measurement update. The multiple EOT problem is addressed using a trajectory Poisson multi-Bernoulli mixture (TPMBM) filter, which is a state-of-the-art method for tracking objects that may generate multiple detections per sensor scan \cite{xia2019extended,xia2023trajectory}. 

In Bayesian filtering, the state of each traffic line is modeled as a uniform B-spline trajectory, parameterized by a sequence of control points. Compared with models based on polylines or clothoids, B-splines can offer greater flexibility and smoothness, making them well-suited for modeling traffic lines with varying curvature \cite{betntorp2023bayesian}. The B-spline trajectory representations also integrate smoothly with the sets of trajectories framework \cite{garcia2019multiple}, and a Bernoulli object spawning model can be easily incorporated to handle lane splitting \cite{garcia2022tracking}. Furthermore, by leveraging the road orientation information provided by the SD map in the lane marking measurement model, we can achieve recursive B-spline estimation \cite{li2023embedding} under Gaussian approximations using a linear Kalman filter.

\subsection{Related Work}

In many existing works, e.g., \cite{abramov2016multi,kim2018crowd,liebner2019crowdsourced,cudrano2022clothoid}, the problem of lane-level map generation using only onboard sensors is typically addressed using graph-based optimization, whose performance can be highly dependent on the quality of data associations of lane marking detection points. However, the data association problem in these works is only addressed using simple heuristics, such as the nearest neighbor approach, which may lead to inaccurate estimates in some challenging cases, e.g., when the vehicle is moving on roads with multiple parallel dashed lane markings. In contrast, our proposed solution uses a Bayesian SLAMOT filter, which can handle the data association problem in a principled way by leveraging data association techniques developed for multiple EOT. 

Multi-lane estimation has also been formulated as a multiple EOT problem in \cite{lundquist2010extended} with polyline lane shape modeling and in \cite{akbari2022tracking} with Gaussian process lane shape modeling. However, these works only consider multi-lane estimation on a frame-by-frame basis, and they do not provide continuous, multi-lane estimates over time, and thus cannot generate lane-level maps. The modeling of each traffic line as a spline trajectory is also considered in \cite{zhao2012novel,qiao2023online} using Catmull-Rom splines. Compared to B-splines, Catmull-Rom splines do not guarantee curvature continuity, which means that the transition between segments may not be smooth.

\subsection{Contributions}
Our proposed method stands out from the existing works by providing a fully Bayesian solution to generate lane-level maps using onboard sensors. Importantly, the multi-trajectory posterior density captures the uncertainties in both the number of traffic lines and their geometries, which are important for fusing the maps generated from multiple vehicle traversals.

The main contributions of this paper are as follows:
\begin{enumerate}
    \item We present a TPMBM-based SLAMOT filter and apply it to the problem of simultaneous localization and multi-lane tracking using onboard sensors and a SD map.
    \item We integrate recursive B-spline estimation into the extended object TPMBM filter to estimate the 3D geometry of traffic lines using lane marking detection points.
    \item The efficacy of proposed solution has been verified using real-world data.
\end{enumerate}

The rest of the paper is organized as follows. In Section~\ref{sec_system_model}, we describe the system model. We present the problem formulation in Section \ref{sec_problem_formulation} and the TPMBM-based SLAMOT filter in Section \ref{sec_tpmbm_based_slmlt}. In Section \ref{sec_experimental_results}, we present the experimental results. Finally, we conclude the paper in Section \ref{sec_conclusion}.

\section{System Model}\label{sec_system_model}

In this section, we introduce the system modeling, which includes the SD map, the state representations of vehicle and traffic lines, as well as their dynamic and measurement models. We assume that the vision data has already been transformed from the camera coordinate to the vehicle coordinate. States and measurements in the vehicle coordinate are denoted with a superscript $l$, and, if not otherwise specified, they are given in an East-North-Up coordinate relative to a local origin.

We use $k\in \mathbb{N}_0$ to denote discrete time steps, and the time instant at time step $k$ is $t_k$, with uniform time interval $\tau = t_{k+1}-t_k$. The cardinality of a set $\mathbf{x}$ is denoted $|\mathbf{x}|$. We use $\uplus$ to denote the union of sets that are mutually disjoint and $\otimes$ to denote the Kronecker product. In addition, we use $\delta_{x}[\cdot]$ and $\delta_{x}(\cdot)$ to represent the Kronecker delta and the Dirac delta functions, respectively, centered at $x$.

\subsection{SD Map}

The SD map can be represented as a directed graph $\mathcal{G} = (\mathcal{V},\mathcal{E})$, where $\mathcal{V}$ is the set of nodes and $\mathcal{E}$ is the set of edges. Each node $v_i \in \mathcal{V}$ is associated with a 2D position $[x_i,y_i]^T\in \mathbb{R}^2$, and each edge $e_{i,j} \in \mathcal{E}$, with start and end nodes $v_i$ and $v_j$, represents a street segment with direction from $v_i$ to $v_j$. Each edge $e_{i,j}$ is described by its length $l_{e_{i,j}}$ and its orientation $\phi_{e_{i,j}}$ via $l_{e_{i,j}} = \|(x_j-x_i,y_j-y_i)\|_2$ and $\phi_{e_{i,j}} = {\displaystyle \arctan}\left(\frac{y_j-y_i}{x_j-x_i}\right)$.
As in \cite{brubaker2015map}, all the street segments are assumed to be one-way, and two-way streets are converted to two one-way streets. 

\subsection{Vehicle State}

The vehicle's 6D pose $p_k = [x_k,y_k,z_k,\alpha_k,\beta_k,\gamma_k]^T$ at time step $k$ consists of its 3D position $[x_k,y_k,z_k]^T \in \mathbb{R}^3$, and its yaw $\alpha_k$, pitch $\beta_k$, and roll $\gamma_k$. The position of the vehicle $q_k = [e_k,d_k]^T$ within a SD map is defined by the street segment $e_k\in \mathcal{E}$ that the vehicle is moving on and the distance $d_k$ from the start node of $e_k$. The inclusion of $d_k$ to the vehicle's state representation may seem redundant, but as we shall see later, it is helpful for modeling the transition of $e_k$. Also note that, as the geometrical information provided by SD map may be inaccurate, the distance $d_k$ may differ from the actual distance between the vehicle's position (projected onto $e_k$) and the start node of $e_k$. The complete vehicle state $o_k = [p_k^T,q_k^T]^T$ at time step $k$ is then 8D, and it consists of the vehicle's global 6D pose $p_k$ and its location $q_k$ within a SD map $\mathcal{G}$.

\subsubsection{Vehicle Measurement Model}
\label{subsubsection_vehicle_measurement_model}

In this work, we assume that the GNSS only measures\footnote{The GNSS operates at a lower frequency than the camera. We assume that the GNSS data has been interpolated to be synchronized with the vision data. It is also possible to directly work with low-frequency GNSS data by considering continuous-discrete particle filtering.} the vehicle's 3D position, and the measurement at time step $k$ is denoted as $\zeta_k$. Assuming that the measurement noise is zero-mean Gaussian, the measurement model of GNSS is given by
\begin{equation}
    \ell_k(\zeta_k|p_k) = \mathcal{N}\left(\zeta_k;[x_k,y_k,z_k]^T,\Omega_k\right),
\end{equation}
where $\Omega_k$ is the measurement noise covariance.

In addition to GNSS, the SD map $\mathcal{G}$ can also be regarded as a measurement source, in the sense that the vehicle's yaw $\alpha_k$ should not diverge too much from the orientation of the street segment $e_k$ that the vehicle is moving on. This measurement likelihood is specified by a von Mises distribution \cite{merriaux2015fast}
\begin{equation}
    \ell_k(\mathcal{G}|o_k) = \mathcal{VM}(\phi_{e_k}|\alpha_k,\kappa),
\end{equation}
where $\mathcal{VM}(\cdot|\mu,\kappa)$ denotes a von Mises distribution with mean $\mu$ and concentration parameter $\kappa$.

\subsubsection{Vehicle Motion Model}\label{subsubsection_vehicle_motion_model}

In this work, we assume that the vehicle motion estimates obtained from the visual odometry can be used as control input\footnote{Vehicle odometry from wheel encoders is used if visual odometry is not available, e.g., when the light condition is poor.}, and that the SD map does not impose any constraints on the vehicle's motion. The vehicle state transition density can then be factorized as
\begin{multline}\label{eq_vehicle_motion_model}
    g_{k+1}\left(p_{k+1},q_{k+1}|p_k,q_k,u^l_{k+1},\mathcal{G}\right) \\ = g^p_{k+1}\left(p_{k+1}|p_k,u^l_{k+1}\right) g^q_{k+1}(q_{k+1}|p_{k+1},q_k,\mathcal{G}),
\end{multline}
where $u^l_{k+1}$ represents the control input in the vehicle coordinate at time step $k+1$, and it can be transformed to the global coordinate $u_{k+1}$ using the vehicle's pose $p_k$.
Assuming that the process noise is zero-mean Gaussian, the transition density of the vehicle's pose at time step $k+1$ is given by 
\begin{equation}
    g^p_{k+1}\left(p_{k+1}|p_k,u^l_{k+1}\right) = \mathcal{N}(p_{k+1};p_k+u_{k+1},Q_{k+1}),
\end{equation}
where $Q_{k+1}$ is the motion noise covariance.

The transition density $g^q_{k+1}(\cdot|p_{k+1},q_k,\mathcal{G})$ of the vehicle's position within a SD map needs to consider the following two cases depending on the distance the vehicle has moved on the street segment: 1) the vehicle stays on the same street segment, and 2) the vehicle moves to a new street segment. In the first case, we have $e_{k+1} = e_k$, and the vehicle's predicted distance to the start node of $e_{k}$ is modeled as $d_{k+1} = d_k + \epsilon_{k+1} + \sigma_{k+1}$, where $\epsilon_{k+1}$ is the vehicle's moved distance along the street segment $e_{k}$, given by 
\begin{equation}\label{eq_vehicle_moved_distance}
    \epsilon_{k+1} = \cos(\phi_{e_k})(x_{k+1}-x_k) + \sin(\phi_{e_k})(y_{k+1}-y_k),
\end{equation}
and $\sigma_{k+1}$ is a zero-mean Gaussian noise. Then for $d_k + \epsilon_{k+1} \leq l_{e_k}$, we have
\begin{equation}
    g^q_{k+1}(q_{k+1}|p_{k+1},q_k,\mathcal{G}) = \delta_{e_k}[e_{k+1}]\mathcal{N}(d_{k+1};d_k+\epsilon_{k+1},\sigma_{k+1}).
\end{equation}

In the second case, we have $e_{k+1} \in N(e_k)$, where $N(e_k)$ is the set of street segments to which $e_k$ connects. Following \cite{merriaux2015fast}, we assume that the difference between the orientation of the new street segment $e_{k+1}$ and the vehicle's predicted yaw $\alpha_{k+1}$ follows a zero-mean von Mises distribution. That is, for each $e \in N(e_k)$,
\begin{equation}\label{eq_vehicle_orientation_transition}
    P(e_{k+1} = e) \propto \mathcal{VM}(\phi_{e}|\alpha_{k+1},\kappa).
\end{equation}
After moving to $e_{k+1}$, the vehicle's distance $d_{k+1}$ to the start node of $e_{k+1}$ needs to have $l_{e_k}$ subtracted, and it is modeled as $d_{k+1} = d_k + \epsilon_{k+1} - l_{e_k} + \sigma_{k+1}$ \cite{brubaker2015map}. Then for $d_k + \epsilon_{k+1} > l_{e_k}$, $e \in N(e_k)$, we have 
\begin{multline}
    g^q_{k+1}(q_{k+1}|p_{k+1},q_k,\mathcal{G}) \\= P(e_{k+1} = e)\mathcal{N}(d_{k+1};d_k+\epsilon_{k+1}-l_{e_k},\sigma_{k+1}).
\end{multline}

\subsection{Traffic Line State}

A traffic line is modeled as a B-spline with uniform knots. A B-spline of degree $d$ has minimum $d+1$ control points and continuous time derivatives up to the $(d-1)$-th order. In this work, we consider 3D quadratic B-splines with $d=2$, which we find sufficient for modeling lane geometry. 

A B-spline trajectory can be parameterized by $C = (\varepsilon,c^{1:\nu})$, where $\varepsilon$ is the initial time step of the trajectory $C$, $\nu \geq 3$ is its length, and $c^{1:\nu} = (c^1,\dots,c^\nu)$, with $c^i \in \mathbb{R}^3$, denotes a finite sequence of control points. In the sequence $c^{1:\nu}$, the first three control points $c^{1:3}$ correspond to time instant $t^c_\varepsilon = t_\varepsilon$, and the rests $c^{4:\nu}$ are at time instants $t^c_{\varepsilon+3},\dots,t^c_{\varepsilon+\nu-1}$, with  uniform interval $\tau^c \geq \tau$. Note that different from the trajectory representation in \cite{garcia2019multiple}, here each trajectory has minimum length 3, and the time interval $\tau^c$ for placing the control points can be larger than the measurement sampling interval $\tau$ \cite{li2023embedding}. 

The trajectory in continuous time can be obtained by interpolating the control points using the B-spline basis function. Specifically, given $C = (\varepsilon,c^{1:\nu})$, the position of a point on the trajectory at an arbitrary time instant $\varsigma \in [t^c_{\varepsilon+i-1},t^c_{\varepsilon+i}), i\in\{1,\dots,\nu-2\}$, is determined by the subsequence of control points $c^{i:i+2}$, according to 
\begin{subequations}\label{eq_b_spline_interpolation}
\begin{align}
    c^\varsigma &= \begin{bmatrix}
        c^i,c^{i+1},c^{i+2}
    \end{bmatrix} \cdot \Sigma \cdot \iota(\varsigma,t^c_{\varepsilon+i-1}), \\
    \iota(\varsigma,t^c_{\varepsilon+i-1}) &= \begin{bmatrix}
        1, \frac{\varsigma-t^c_{\varepsilon+i-1}}{\tau^c}, \left(\frac{\varsigma-t^c_{\varepsilon+i-1}}{\tau^c}\right)^2
    \end{bmatrix}^T, \\
    \Sigma &= \frac{1}{2}\begin{bmatrix}
        1 & -2 & 1 \\
        1 & 2 & -2 \\
        0 & 0 & 1
    \end{bmatrix}.
\end{align}
\end{subequations}
where $\Sigma$ is the basis function matrix for quadratic B-splines.

We consider B-spline trajectories up to the current time step $k$. Trajectory $C = (\varepsilon,c^{1:\nu})$ is considered alive at time step $k$ if and only if the latest measurement time instant $t_k$ satisfies that $t_k \in [t^c_{\varepsilon+\nu-3},t^c_{\varepsilon+\nu-2})$, i.e., $\nu - \lfloor \frac{\tau(k-\varepsilon)}{\tau^c}\rfloor = 3$. In the rest of the paper, we use the notation $\lfloor \frac{\tau(k-\varepsilon)}{\tau^c}\rfloor \triangleq \Xi^{\tau,\tau^c}_{k,\varepsilon}$ for brevity, where $\Xi^{\tau,\tau^c}_{k,\varepsilon}+3$ represents the maximum length of trajectory $C = (\varepsilon,c^{1:\nu})$ at time step $k$. Then, the space of a single trajectory up to time step $k$ can be defined as $T_{(k)} = \uplus_{(\varepsilon,\nu) \in I_{(k)}} \{\varepsilon\} \times \mathbb{R}^{3\nu}$, where $I_{(k)} = \{(\varepsilon,\nu): 0 \leq \varepsilon \leq k \text{ and } 3 \leq \nu \leq 3 + \Xi^{\tau,\tau^c}_{k,\varepsilon}\}$. The set of all trajectories that have existed up to time step $k$ is denoted as $\mathbf{C}_k$. Definitions of integrals and densities of single trajectories and sets of trajectories can be found in \cite{garcia2019multiple}.

\subsubsection{Multi-Lane Measurement Model}
\label{subsubsection_multi_lane_measurement_model}

From \eqref{eq_b_spline_interpolation}, we can see that, for an alive trajectory $C = (\varepsilon,c^{1:\nu})$ at time step $k$, its interpolation at time step $k$ is determined by its latest three control points $c^{\nu-2:\nu}$. This means that the detections of $C$ at time step $k$ only depend on $c^{\nu-2:\nu}$ (and the vehicle state).

A single B-spline trajectory $C = (\varepsilon,c^{1:\nu})$ at time step $k$ is detected with probability
\begin{equation}
    P_k^D(C,p_k) = \begin{cases}
        P_k^D\left(c^{\nu-2:\nu},p_k\right), & \nu - \Xi^{\tau,\tau^c}_{k,\varepsilon} = 3, \\
        0, & \text{otherwise},
    \end{cases}
\end{equation}
and generates an independent set $\mathbf{w}^{L,l}_k\in \mathbb{R}^3$ of lane marking edge detection points in the vehicle coordinate, or is misdetected with probability $1-P_k^D(C,p_k)$. For a detection point $w^l_k$, it can be transformed to the global coordinate $w_k \in \mathbf{w}^L_k$ using the vehicle's pose $p_k$, and the dependence of $w_k$ on $p_k$ is made implicit in the following expressions.

% The set measurement likelihood of $\mathbf{w}^L_k$ can be expressed as
% \begin{multline}
%     \label{eq_lane_measurement_model}
%     \ell_k \left(\mathbf{w}^L_k|\varepsilon,c^{\nu-2:\nu}\right) = \\ \begin{cases}
%         \begin{aligned}
%             &P_k^D(C,p_k)\left|\mathbf{w}^L_k\right|!P\left(\left|\mathbf{w}^L_k\right| | c^{\nu-2:\nu}\right)\\
%             &\times \prod_{w_k \in \mathbf{w}^L_k} \ell_k \left(w_k|\varepsilon,c^{\nu-2:\nu}\right)
%         \end{aligned} & \mathbf{w}^L_k \neq \emptyset, \\
%         1-P_k^D(C,p_k) + P\left(\left|\mathbf{w}^L_k\right| = 0| c^{\nu-2:\nu}\right)& \mathbf{w}^L_k = \emptyset,
%     \end{cases}
% \end{multline}
% where $\ell_k (w_k|\varepsilon,c^{\nu-2:\nu})$ is the single measurement likelihood. In the second case, $P(|\mathbf{w}^L_k| = 0| c^{\nu-2:\nu})$ accounts for the case that trajectory $C$ is detected but generates zero measurement. 

We assume that each individual lane marking edge detection point $w_k$ originates from a measurement source $\varpi_k$, corrupted by a zero-mean Gaussian noise with covariance $\Omega_k^{w_k}$. Each measurement source $\varpi_k$ is considered uniformly distributed along the two lane marking edges. This modeling assumption allows us to rewrite the single measurement likelihood as
\begin{equation}\label{eq_single_measurement_likelihood}
    \ell_k \left(w_k|\varepsilon,c^{\nu-2:\nu}\right) = \int \mathcal{N}\left(w_k;\varpi_k,\Omega_k^{w_k}\right)\mathcal{U}\left(\varpi_k|\varepsilon,c^{\nu-2:\nu}\right) d\varpi_k.
\end{equation}
To enable the tractable computation of \eqref{eq_single_measurement_likelihood}, we further approximate the uniform distribution $\mathcal{U}(\varpi_k|\varepsilon,c^{\nu-2:\nu})$ as a Gaussian $\mathcal{N}(\varpi_k;h(\varepsilon,c^{\nu-2:\nu}),\Omega_k^\varpi)$, which yields
\begin{equation}
    \ell_k \left(w_k|\varepsilon,c^{\nu-2:\nu}\right) = \mathcal{N}\left(w_k;h\left(\varepsilon,c^{\nu-2:\nu}\right),\Omega_k^{w_k} + \Omega_k^\varpi\right).
\end{equation}

In Gaussian distribution  $\mathcal{N}(\varpi_k;h(\varepsilon,c^{\nu-2:\nu}),\Omega_k^\varpi)$, the mean $h(\varepsilon,c^{\nu-2:\nu})$ is the interpolated point on trajectory $C$ at time step $k$, and it can be computed using \eqref{eq_b_spline_interpolation}, which gives
\begin{equation}
    h(\varepsilon,c^{\nu-2:\nu}) = \left(\Sigma \cdot \iota(t_k,t^c_{\varepsilon+\nu-3})\right)^T \otimes I_3 c^{\nu-2:\nu},
\end{equation}
where $I_3$ is an identity matrix. The covariance matrix $\Omega_k^\varpi$ can be factorized as \cite{yang2019tracking}
\begin{equation}\label{eq_measurement_source_covariance}
    \Omega_k^\varpi = R_k^\varpi \text{diag}\left([l_x,l_y,0]/2\right)  \cdot \Lambda \cdot \text{diag}\left([l_x,l_y,0]/2\right) {R_k^\varpi}^T,
\end{equation}
where $R_k^\varpi$ is a 3D rotation matrix that aligns the measurement source $\varpi_k$ with the orientation of the lane marking, and $\Lambda$ is a diagonal matrix, which can be set to $\Lambda = I_3/4$ to match the Gaussian distribution to a uniform distribution \cite{yang2019tracking}. The factorization \eqref{eq_measurement_source_covariance} is motivated by the fact that each observed lane marking within a  certain distance of the vehicle can be approximated as a very thin rectangle with length $l_x$ and width $l_y$. The length $l_x$ setting should consider the trade-off between detection range and approximation error due to road curvature, whereas the width $l_y$ can be determined by lane marking type. 

% For example, if we consider lane marking detection points that are within $[x_{min}~\text{m},x_{max}~\text{m}]$ from the vehicle in its $x$-axis, we can set $l_x = (x_{max} - x_{min})~\text{m}$.

The rotation matrix $R_k^\varpi$ in \eqref{eq_measurement_source_covariance} depends on the orientation of the (observed) lane marking at time step $k$, which can be approximately obtained by computing the tangent vector of the velocity of trajectory $C$ (the first derivative of \eqref{eq_b_spline_interpolation} with respect to time) at time step $k$. However, in this case the measurement noise covariance $\Omega_k^\varpi$  would become state-dependent, resulting in a (highly) nonlinear measurement model \cite{spinello2010nonlinear}. 

To compute $R_k^\varpi$ in a simple yet effective way, we assume that the yaw of the B-spline trajectory evaluated at the interpolated point $h(\varepsilon,c^{\nu-2:\nu})$ is given by the orientation $\phi_{e^\prime}$ of the street segment $e^\prime$ that $h(\varepsilon,c^{\nu-2:\nu})$ is on, and that the pitch and roll of the B-spline trajectory evaluated at $h(\varepsilon,c^{\nu-2:\nu})$ align with the vehicle's. Under these assumptions, we have that 
\begin{equation}
    R_k^\varpi = R(\phi_{e^\prime},\beta_k,\gamma_k),
\end{equation}
where $R(\cdot)$ is a 3D rotation matrix determined by yaw, pitch, and roll. This assumption is reasonable for typical cases where the lane markings are in parallel with the street segment and the street segments are flat within a short range of distance. To find $e^\prime$, we first compute the distance $\epsilon^\prime$ between the vehicle's position and the interpolated point $h(\varepsilon,c^{\nu-2:\nu})$ projected on the street segment $e_k$ that the vehicle is currently moving on, similar to \eqref{eq_vehicle_moved_distance}. If $d_k + \epsilon^\prime \leq l_{e_k}$, we have $e^\prime = e_k$. Otherwise, we find the street segment $e 
\in N(e_k)$ that maximizes \eqref{eq_vehicle_orientation_transition} and check if $d_k + \epsilon^\prime - l_{e_k} \leq l_{e}$. This process is repeated until we find $e^\prime$. In addition to the detection points generated by lane markings, we assume that the lane marking detector can also report clutter detection points. The set $\mathbf{w}_k^{C,l}$ of clutter detection points is modeled as a Poisson point process (PPP) with Poisson clutter rate $\lambda^C_k$. The set of all detection points at time step $k$ is denoted as $\mathbf{w}^l_k$.

\subsubsection{Multi-Lane Dynamic Model}\label{subsubsection_multi_lane_dynamic_model}

Given the set of all trajectories $\mathbf{C}_k$ at time step $k$, each $C=(\varepsilon,c^{1:\nu}) \in \mathbf{C}_k$ survives to time step $k+1$, i.e., remains in the set $\mathbf{C}_{k+1}$, with probability one, and its transition density is given by
\begin{multline}\label{eq_lane_dynamic_model}
    g_{k+1}\left(\varepsilon_+,c_+^{1:\nu_+} | C\right) = \delta_\varepsilon[\varepsilon_+]\\
    \times \begin{cases}
        \delta_\nu[\nu_+]\delta_{c^{1:\nu}}\left(c_+^{1:\nu_+}\right), & \begin{aligned}
            &\nu < 3 + \Xi^{\tau,\tau^c}_{k,\varepsilon} \text{or},\\
            &\nu = 3 + \Xi^{\tau,\tau^c}_{k,\varepsilon}\text{and}\\
            &t_{k+1} < t^c_{\varepsilon+\nu-2},
        \end{aligned}\\
        \begin{aligned}
            &\delta_\nu[\nu_+]\delta_{c^{1:\nu}}\left(c_+^{1:\nu_+}\right)\\
            &\times\left(1-P^S_1(c^\nu)\right),
        \end{aligned}
         & \begin{aligned}
            &\nu = 3 + \Xi^{\tau,\tau^c}_{k,\varepsilon}\text{and} \\ &t_{k+1} \geq t^c_{\varepsilon+\nu-2}\text{ and}\\
            &t_{k+1} \geq t^c_{\varepsilon_+ + \nu_+ -2},
         \end{aligned} \\
         \begin{aligned}
            &\delta_{\nu+1}[\nu_+]\delta_{c^{1:\nu}}\left(c_+^{1:\nu_+-1}\right)\\
            &\times P^S_1(c^\nu)g\left(c_+^{\nu_+}|c^{\nu-1:\nu}\right),
        \end{aligned}
         & \begin{aligned}
            &\nu = 3 + \Xi^{\tau,\tau^c}_{k,\varepsilon}\text{and}\\ &t_{k+1} \geq t^c_{\varepsilon+\nu-2}\text{ and}\\
            &t_{k+1} < t^c_{\varepsilon_+ + \nu_+ -2}.
         \end{aligned}
    \end{cases}
\end{multline}
From \eqref{eq_lane_dynamic_model}, we can see that if trajectory $C$ is not alive at time step $k$, or if it is alive at time step $k$ but the time instant $t_{k+1}$ is smaller than $t^c_{\varepsilon+\nu-2}$, then trajectory $C$ remains unaltered with probability one. If trajectory $C$ is alive at time step $k$ and the time instant $t_{k+1}$ is no smaller than $t^c_{\varepsilon+\nu-2}$, then with probability $1-P^S_1(c^\nu)$, trajectory $C$ remains unaltered, and with probability $P^S_1(c^\nu)$, it is extended by appending one control point with a transition density $g(c_+^{\nu_+}|c^{\nu-1:\nu})$. 

The state of the new control point $c^{\nu_+}$ in $g(c_+^{\nu_+}|c^{\nu-1:\nu})$ is determined by the state of the latest two control points $c^{\nu-1:\nu}$ by preserving the velocity at time instant $t^c_{\varepsilon+\nu-2}$, which gives $c_+^{\nu_+} = 2c^{\nu} - c^{\nu-1}$. Assuming that the motion process noise is zero-mean Gaussian, the transition density is
\begin{equation}\label{eq_transition}
    g\left(c_+^{\nu_+}|c^{\nu-1:\nu}\right) = \mathcal{N}\left(c_+^{\nu_+};\begin{bmatrix}-I_3 & 2I_3\end{bmatrix}c^{\nu-1:\nu},Q_{k+1}^c\right),
\end{equation}
where $Q^c_{k+1}$ is the motion noise covariance.

The set of trajectories $\mathbf{C}_{k+1}$ at time step $k+1$ is the union of surviving trajectories, newborn trajectories, and trajectories spawned from existing trajectories. The set of newborn trajectories appears independently with a PPP with intensity 
\begin{equation}\label{eq_birth_intensity}
    \lambda^B_{k+1}(\varepsilon,c^{1:\nu} | p_{k+1}) = \delta_{k+1}[\varepsilon]\delta_3[\nu]\lambda^B_{k+1}(c^{1:3} | p_{k+1} ),
\end{equation}
where the Poisson intensity $\lambda^B_{k+1}(c^{1:3} | p_{k+1} )$ depends on the camera's field-of-view. To model lane splitting, we assume that each trajectory $C=(\varepsilon,c^{1:\nu}) \in \mathbf{C}_k$ alive at time step $k$ may spawn a new trajectory with probability $P^S_2(c^\nu)$. The set of trajectories spawned from trajectory $C$, where $\nu - \Xi^{\tau,\tau^c}_{k,\varepsilon} = 3$, at time step $k+1$ is then a Bernoulli process with density 
\begin{align}
    &g_{k+1}^S\left(\mathbf{C}_+|C\right)\nonumber \\ &= \begin{cases}
        P^S_2(c^\nu)\delta_{k+1}[\varepsilon_+]\delta_3[\nu_+] &
        \\
        ~~\times \delta_{c^{\nu-1:\nu}}\left(c_+^{1:\nu_+-1}\right) &\\~~\times g\left(c_+^{\nu_+}|c^{\nu-1:\nu}\right), & \mathbf{C}_+ = \left\{\left(\varepsilon_+,c_+^{1:\nu_+}\right)\right\},\\
        1 - P^S_2(c^\nu), & \mathbf{C}_+ = \emptyset, \\
        0, & \text{otherwise}.
    \end{cases}
\end{align}

\section{Problem Formulation}\label{sec_problem_formulation}

With the system model introduced in Section \ref{sec_system_model} in place, we can now formulate the problem of simultaneous localization and multi-lane tracking. The objective is to recursively compute the joint posterior distribution of the vehicle trajectory $o_{0:k}$ and the set $\mathbf{C}_k$ of all B-spline trajectories given the SD map $\mathcal{G}$ and the measurements $\mathbf{w}^l_{1:k}$ and visual odometry $u^l_{1:k}$ up to and including time step $k$. The joint posterior distribution can be factorized as
the product of the posterior distribution $f(o_{0:k}|\zeta_{1:k},u^l_{1:k},\mathcal{G})$ of the vehicle trajectory and the posterior distribution $f(\mathbf{C}_k|o_{0:k},\mathbf{w}^l_{1:k},\mathcal{G})$ of the set of all trajectories conditioned on the vehicle trajectory, i.e.,
\begin{align}\label{eq_joint_posterior_distribution}
    &f\left(o_{0:k},\mathbf{C}_k|\zeta_{1:k},u^l_{1:k},\mathbf{w}^l_{1:k},\mathcal{G}\right) \nonumber
    \\ &=f\left(o_{0:k}|\zeta_{1:k},u^l_{1:k},\mathbf{w}^l_{1:k},\mathcal{G}\right)f\left(\mathbf{C}_k|o_{0:k},\mathbf{w}^l_{1:k},\mathcal{G}\right).
\end{align}

The computation of each posterior density involves a prediction step and an update step.
For the vehicle motion model in Section \ref{subsubsection_vehicle_motion_model}, the predicted density of vehicle trajectory at time step $k$ is 
\begin{align}\label{eq_vehicle_prediction_step}
    &f\left(o_{0:k}|\zeta_{1:k-1},u^l_{1:k},\mathbf{w}^l_{1:k-1},\mathcal{G}\right) = g_{k}\left(o_{k}|o_{k-1},u^l_{k},\mathcal{G}\right)\nonumber\\ &\times f\left(o_{0:k-1}|\zeta_{1:k-1},u^l_{1:k-1},\mathbf{w}^l_{1:k-1},\mathcal{G}\right).
\end{align}
The predicted density of the set of all B-spline trajectories at time step $k$ is
\begin{align}\label{eq_multi_lane_prediction_step}
    &f\left(\mathbf{C}_k|o_{0:k},\mathbf{w}^l_{1:k-1},\mathcal{G}\right) = \nonumber\\ &\int g_k\left(\mathbf{C}_k | \mathbf{C}_{k-1}\right) f\left(\mathbf{C}_{k-1}|o_{0:k},\mathbf{w}^l_{1:k-1},\mathcal{G}\right) \delta \mathbf{C}_{k-1},
\end{align}
where $g_k(\mathbf{C}_k | \mathbf{C}_{k-1}) $ is the transition density of the set of all trajectories for the multi-lane dynamic model in Section \ref{subsubsection_multi_lane_dynamic_model}, and the set integral $\int f(\mathbf{X})\delta \mathbf{X}$ is defined in \cite{garcia2019multiple}.

The predicted density of the set of all B-spline trajectories at time step $k$ is then updated using the multi-lane measurement model $\ell_k(\mathbf{w}_k^l | \mathbf{C}_k,o_k,\mathcal{G})$ in Section \ref{subsubsection_multi_lane_measurement_model}, which gives
\begin{align}\label{eq_multi_lane_update_step}
    &f\left(\mathbf{C}_k|o_{0:k},\mathbf{w}^l_{1:k},\mathcal{G}\right)\nonumber\\ &=\frac{f\left(\mathbf{C}_k|o_{0:k},\mathbf{w}^l_{1:k-1},\mathcal{G}\right) \ell_k\left(\mathbf{w}_k^l | \mathbf{C}_k,o_k,\mathcal{G}\right)}{\ell_k\left(\mathbf{w}_k^l | o_{0:k}, \mathbf{w}_{1:k-1}^l,\mathcal{G}\right)},
\end{align}
where $\ell_k(\mathbf{w}_k^l | o_{0:k}, \mathbf{w}_{1:k-1}^l, \mathcal{G})$ is a normalizing factor. Lastly, the predicted density of the vehicle trajectory at time step $k$ is updated using the measurement model in Section \ref{subsubsection_vehicle_measurement_model}, which gives
\begin{align}\label{eq_vehicle_update_step}
    &f\left(o_{0:k}|\zeta_{1:k},u^l_{1:k},\mathbf{w}^l_{1:k},\mathcal{G}\right) = f\left(o_{0:k}|\zeta_{1:k-1},u^l_{1:k},\mathbf{w}^l_{1:k-1},\mathcal{G}\right) \nonumber \\
    &~~~\times \ell_k\left(\zeta_k,\mathcal{G}|o_{k}\right) \ell_k\left(\mathbf{w}_k^l | o_{0:k}, \mathbf{w}_{1:k-1}^l,\mathcal{G}\right) \nonumber \\
    &~~~~\slash \ell_k\left(\zeta_k,\mathbf{w}^l_{k},\mathcal{G}|\zeta_{1:k-1},\mathbf{w}^l_{1:k-1}\right),
\end{align}
where $\ell_k(\zeta_k,\mathbf{w}^l_{k},\mathcal{G}|\zeta_{1:k-1},\mathbf{w}^l_{1:k-1})$ is a normalizing factor.

In the next section, we will describe how to perform these prediction and update steps in a computationally tractable way. For notational simplicity, we will drop the explicit dependence in the predicted and posterior densities and use the shorthand notations, with $k^\prime \in \{k-1,k\}$,
\begin{subequations}
    \begin{align}
        f\left(o_{0:k}|\zeta_{1:k^\prime},u^l_{1:k},\mathbf{w}^l_{1:k^\prime},\mathcal{G}\right) &= f_{k|k^\prime}\left(o_{0:k}\right), \\
        f\left(\mathbf{C}_k|o_{0:k},\mathbf{w}^l_{1:k^\prime},\mathcal{G}\right) &= f_{k|k^\prime}\left(\mathbf{C}_k| o_{0:k}\right).
    \end{align}
\end{subequations}

\section{TPMBM-based Simultaneous Localization and Multi-Lane Tracking}\label{sec_tpmbm_based_slmlt}

The main challenge in computing \eqref{eq_joint_posterior_distribution} lies in the multi-lane update step \eqref{eq_multi_lane_update_step}, which involves the dependence on the vehicle trajectory $o_{0:k}$, and the normalizing factor in \eqref{eq_multi_lane_update_step} needs to be explicitly computed for updating the vehicle trajectory in \eqref{eq_vehicle_update_step}. To address this challenge, we adopt a particle representation of the vehicle trajectory \cite{kim2022pmbm,montemerlo2002fastslam}:
\begin{equation}\label{eq_particle_representation}
    f_{k|k^\prime}\left(o_{0:k}\right) = \sum_{n=1}^{N_p} \omega_{k | k^\prime}^n \delta_{o_{0:k}^n}\left(o_{0:k}\right),
\end{equation}
where $N_p$ is the number of particles and $\omega_{k | k^\prime}^n$ is the weight of particle $o_{0:k}^n$, with $\omega_{k | k^\prime}^n \geq 0$ and $\sum_{n=1}^{N_p}\omega_{k | k^\prime}^n = 1$. 

With this particle representation, the multi-lane prediction \eqref{eq_multi_lane_prediction_step} and update step \eqref{eq_multi_lane_update_step} boil down to computing the predicted and posterior density of the set $\mathbf{C}_k$ of all trajectories for each particle $o_{0:k}^n$. In particular, given the multi-lane measurement and dynamic models in Section \ref{subsubsection_multi_lane_measurement_model} and \ref{subsubsection_multi_lane_dynamic_model}, the density $f_{k|k^\prime}(\mathbf{C}_k | o^n_{0:k})$ is of the form TPMBM with \cite{xia2023trajectory},

\small
\begin{align}
    f_{k|k^\prime}(\mathbf{C}_k | o^n_{0:k}) &= \sum_{\mathbf{X} \uplus \mathbf{Y} = \mathbf{C}_k} f^p_{k|k^\prime}(\mathbf{X} | o^n_{0:k})f^{mbm}_{k|k^\prime}(\mathbf{Y} | o^n_{0:k}), \label{eq_pmbm} \\
    f^p_{k|k^\prime}(\mathbf{C}_k | o^n_{0:k}) &= e^{-\int \lambda_{k|k^\prime}(C | o^n_{0:k}) d C}\prod_{C \in \mathbf{C}_k}\lambda_{k|k^\prime}(C | o^n_{0:k}), \label{eq_ppp}\\
    f^{mbm}_{k|k^\prime}(\mathbf{C}_k | o^n_{0:k}) &= 
    \sum_{a \in \mathcal{A}^n_{k | k^\prime}} \omega^a_{k | k^\prime}  \sum_{\uplus_{l=1}^{I^n_{k | k^\prime}}\mathbf{C}^l = \mathbf{C}_k} \prod_{i=1}^{I^n_{k|k^\prime}} f^{i,a^i}_{k | k^\prime}\left(\mathbf{C}^i | o^n_{0:k}\right).\label{eq_mbm}
\end{align}
\normalsize
The TPMBM in \eqref{eq_pmbm} is the union of two independent sets: a trajectory PPP with density \eqref{eq_ppp} that represents undetected trajectories that are hypothesized to exist but have never been detected, and a trajectory multi-Bernoulli mixture with density \eqref{eq_mbm} that represents trajectories that have been detected at some point up to time step $k$. The intensity of the trajectory PPP is $\lambda_{k|k^\prime}(\cdot)$. Each received measurement generates a Bernoulli component, and the number of Bernoulli components is $I^n_{k|k^\prime}$. A global hypothesis is $a = (a^1,\dots,a^{I^n_{k|k^\prime}}) \in \mathcal{A}^n_{k | k^\prime}$, where $a^i\in\{1,\dots,h^i_{k|k^\prime}\}$ is the index to the local hypothesis for the $i$-th Bernoulli and $h^i_{k|k^\prime}$ is its number of local hypotheses. 

We refer to the $j$-th measurement $w_k^{l,j}$ using the pair $(k,j)$, and the set of all such measurement pairs up to (and including) time step $k$ is denoted by $\mathcal{M}_k$. Then, a local hypothesis $a^i$ for the $i$-th Bernoulli component has a set of measurement pairs denoted as $\mathcal{M}_k^{i,a^i} \subseteq \mathcal{M}_k$. The set $\mathcal{A}^n_{k|k^\prime}$ of global hypotheses satisfies \cite{xia2023trajectory}, 
\begin{multline}
    \mathcal{A}^n_{k|k^\prime} = \left\{ \left(a^1,\dots,a^{I^n_{k|k^\prime}}\right): a^i\in\left\{1,\dots,h^i_{k|k^\prime}\right\} \forall i,\right.\\ \left.\uplus_{i=1}^{I^n_{k|k^\prime}} \mathcal{M}_{k^\prime}^{i,a^i} = \mathcal{M}_{k^\prime}\right\}.
\end{multline}
Global hypothesis $a$ has weight $\omega^a_{k | k^\prime} \propto \prod_{i=1}^{I^n_{k|k^\prime}} \omega^{i,a^i}_{k | k^\prime}$, where $\omega_{k|k^\prime}^{i,a^i}$ is the weight of local hypothesis $a^i$ for the $i$-th Bernoulli component, and it should satisfy that $\sum_{a\in\mathcal{A}^n_{k|k^\prime}}\omega_{k|k^\prime}^{a} = 1$. The density of the $i$-th Bernoulli component with local hypothesis $a^i$ is $f^{i,a^i}_{k|k^\prime}(\cdot|o_{0:k}^n)$, parameterized by a probability of existence $r^{i,a^i}_{k|k^\prime}$ and a single trajectory density $p^{i,a^i}_{k|k^\prime}(\cdot|o_{0:k}^n)$.

We proceed to describe the prediction \eqref{eq_vehicle_prediction_step}, \eqref{eq_multi_lane_prediction_step} and update \eqref{eq_multi_lane_update_step}, \eqref{eq_vehicle_update_step}  steps with the particle representation of the vehicle trajectory \eqref{eq_particle_representation} and the TPMBM density representation of the set of all trajectories conditioned on each particle \eqref{eq_pmbm}. Due to page limits, we refer the readers to \cite{xia2023trajectory,garcia2022tracking} for the explicit expressions of the TPMBM prediction and update equations.

\subsection{Vehicle State Prediction}

Given a trajectory particle $o_{0:k-1}^n$, we first draw a sample of the vehicle pose at time step $k$ via $p^n_{k} \sim g_{k}(p_{k}|p^n_{k-1},u^l_{k})$, and then we draw a sample of the vehicle's position within the SD map at time step $k$ via $q_{k}^n \sim g_{k}(q_{k}|p^n_{k},q^n_{k-1},\mathcal{G})$. Lastly, the sampled vehicle state $o^n_k$ is appended to $o_{0:k-1}^n$ to obtain the updated particle $o_{0:k}^n$. Since we directly draw samples from the vehicle motion transition density, the particle weights remain unchanged. 

\subsection{Multi-Lane Prediction}

Given a TPMBM posterior density at time step $k-1$ and the multi-lane dynamic model in Section \ref{subsubsection_multi_lane_dynamic_model} without spawning, the predicted density would also be a TPMBM \cite{xia2023trajectory}. However, due to the dependencies between each alive and its spawned trajectory, the predicted density is no longer a TPMBM in its closed form\footnote{For the standard dynamic and measurement models with multi-Bernoulli spawning, the posterior is a TPMBM density on the set of \textit{tree} trajectories without approximations \cite{garcia2022tracking}. This modeling is not used in this work.}. As in \cite{garcia2022tracking}, we approximate the predicted density as a TPMBM by discarding these dependencies. In addition, we assume that the PPP represents alive trajectories without spawning, and that each trajectory Bernoulli component has a deterministic start time. These approximations would considerably simplify the implementation of the multi-lane prediction step. Note that the TPMBM filter in \cite{garcia2022tracking} considers the sets of tree trajectories, which contain the genealogy information of each trajectory. However, in multi-lane tracking the genealogy information is not needed, and we only consider the traditional sets of trajectories formulation for notational simplicity \cite{garcia2019multiple}.

The prediction of PPP and existing Bernoulli components are the same as their predictions in the TPMBM filter without spawning \cite[Prop. 1]{xia2023trajectory}. For a spawned Bernoulli component, its probability of existence is jointly determined by the spawning probability $P^S_2(\cdot)$, the probability of existence and the probability of being alive at time step $k$ of its parent trajectory. After the prediction step, the number of Bernoulli components \eqref{eq_pmbm} becomes $I^n_{k|k-1} = 2I^n_{k-1|k-1}$.

\subsection{Multi-Lane Update}

Given a TPMBM predicted density at time step $k$ and the multi-lane measurement model in Section \ref{subsubsection_multi_lane_measurement_model}, the posterior density at time step $k$ is also a TPMBM \eqref{eq_pmbm} \cite[Prop. 2]{xia2023trajectory}. The complete TPMBM update consists of five steps: 
\begin{enumerate}
    \item The update of the PPP for undetected trajectories that remain undetected. 
    \item The update of the PPP for undetected trajectories that are detected for the first time. In this step, each received measurement generates a unique Bernoulli component, such that $I^n_{k|k} = I^n_{k|k-1} + |\mathbf{w}_k^l|$.
    \item The update of local hypotheses corresponding to misdetection of existing Bernoulli components.
    \item The update of local hypotheses corresponding to measurement update of existing Bernoulli components. 
    \item The update of global hypotheses.
\end{enumerate}

The main challenge lies in the update of global hypotheses. Each global hypothesis corresponds to a unique partition of $\mathcal{M}_k$, and the number of global hypotheses is given by the Bell number of $|\mathcal{M}_k|$ \cite{pmbmextended2}. Thus, it is computationally intractable to enumerate all the global hypotheses. To address this challenge, we can only consider global hypotheses with high likelihoods, using approaches that combine clustering and 2D assignments \cite{pmbmextended2} or sampling-based methods \cite{soextended}.

% To further reduce the computational complexity, we prune global hypotheses with small weights and Bernoulli components with small existence probability. In addition, Bernoulli components with small probability of being alive at the current time step are considered dead, and they are no longer predicted or updated. We also use an $L$-scan implementation \cite{garcia2019trajectory}, where control points before the last $L$ time instants are considered independent and remain unchanged with new measurements.

\subsection{Vehicle State Update}

Computing the particle weight $\omega^n_{k|k}$ requires the computation of $\ell_k(\mathbf{w}_k^l | o^n_{0:k}, \mathbf{w}_{1:k-1}^l, \mathcal{G})$, which is the normalizing factor in computing the TPMBM posterior density \eqref{eq_multi_lane_update_step} at time step $k$. It can be observed that this normalizing factor is the same as the normalizing factor in computing the normalized global hypothesis weight \cite{kim2022pmbm}, i.e.,
\begin{equation}
    \ell_k(\mathbf{w}_k^l | o^n_{0:k}, \mathbf{w}_{1:k-1}^l, \mathcal{G}) = \sum_{a\in\mathcal{A}^n_{k|k}} \prod_{i=1}^{I^n_{k|k}} \omega^{i,a^i}_{k | k}.
\end{equation}
The updated particle weight is then given by
\begin{equation}
    \omega^n_{k|k} \propto \omega^n_{k|k-1}\ell_k\left(\zeta_k|o_{k}\right)\ell_k\left(\mathcal{G}|o_{k}\right) \sum_{a\in\mathcal{A}^n_{k|k}} \prod_{i=1}^{I^n_{k|k}} \omega^{i,a^i}_{k | k}.
\end{equation}
To avoid particle degeneracy, we resample the particles if the number of effective particles is below $N_p/4$.

\section{Experimental Results}\label{sec_experimental_results}

In this section, we present the experimental results of our proposed solution for simultaneous localization and multi-lane tracking using onboard sensors and a SD map on real-world data. The data was collected by a test vehicle during a single traversal on a highway in Gothenburg, Sweden.

The vision data was obtained using a forward-facing, mono camera, operating at 15 Hz. Both the visual odometry and the lane marking edge detections were extracted from the camera images using Zenseact's in-house developed computer vision algorithms. The GNSS measurements were obtained using a single-antenna GPS receiver, operating at 1 Hz; the SD map is from the open-source OpenStreetMap. 

To evaluate the localization performance of our proposed solution, the vehicle pose estimates reported by a high-precision localization system from Oxford Technical Solutions (OxTS) are used as ground truth. Due to the lack of the ground truth HD map, we evaluate the multi-lane estimation performance by overlaying the estimated traffic lines on the Google Earth satellite image.

The parameter setting used in the SLAMOT filter is specified as follows. In the vehicle measurement model, the GNSS measurement noise covariance $\Omega_k$ is a diagonal matrix. The standard deviation of the latitudinal and longitudinal errors are provided by the GNSS, and we set the standard deviation of the altitude error to 3\,m. The concentration parameter $\kappa$ in the von Mises distribution is set to 8. In the vehicle motion model, the motion noise covariance $Q_{k}$ is a diagonal matrix with the standard deviation of the position error set to 0.05\,m and orientation error set to $0.1^\circ$. The standard deviation $\sigma_k$ of motion noise of $d_k$ is set to $0.2\epsilon_k$, where $\epsilon_k$ is defined in \eqref{eq_vehicle_moved_distance}. For B-spline trajectory state modeling, the time interval for placing the control points is $\tau^c = 5\tau = 1/3\,\text{s}$.

We consider lane marking detection points that are within 16\,m distance of the vehicle and set $l_x = 3\,\text{m}$ and $l_y = 0.05\,\text{m}$ in the covariance matrix \eqref{eq_measurement_source_covariance} that represents the lane extent. In addition to the 3D positional information, each lane marking edge detection point also contains the instance segmentation information provided by the lane marking edge detector. In this work, we have utilized the lane marking type information, such as solid or dashed, and the clustering information, i.e., to which lane marking cluster it belongs. Each traffic line is detected with probability $P^D_k = 0.95$ except for the following special case. If a traffic line was associated to a dashed lane marking detection point at time step $k-1$ and is not detected at time step $k$, then its detection probability will be lowered to $P^D_{k} = 0.5$ until it is detected again. This setting helps to connect multiple dashed lane markings into a single traffic line. In addition, by using the clustering information, the extended object data association problem reduces to the point object data association problem. Each individual detection point $w_k$ also has its own $\Omega_k^{w_k}$ provided by the lane marking detector. The clutter detections have uniform Poisson intensity $10^{-3}$.

In the multi-lane dynamic model, the survival probability is $P^S_1 = 0.99$ and the spawning probability is $P^S_2 = 10^{-5}$. The motion noise covariance in the state transition density \eqref{eq_transition} is $Q^c_{k} = 0.36\tau^cI_3$. The Poisson birth intensity \eqref{eq_birth_intensity} is a single Gaussian defined in the vehicle coordinate, with weight $10^{-5}$, mean $[6,0,-0.5,11,0,-0.5,16,0,-0.5]^T$, and covariance $\text{diag}(100,400,4,100,400,4,100,400,4)$. The weight at time step $0$ is set to $10^{-3}$.

% \begin{figure}
%     \centering
%     \includegraphics[width=\columnwidth]{satelliteImage.jpg}
%     \caption{OxTS vehicle trajectory is shown in blue, overlaid with Google Earth satellite image. It represents a single traversal of a test vehicle driving south on a highway in Gothenburg, Sweden.}
%     \label{fig_satellite_image}
% \end{figure}

The TPMBM update uses Murty's algorithm \cite{crouse2016implementing} to find the $M = \lceil 500 \omega^a_{k|k}\rceil$-best global hypotheses for global hypothesis $a$ with weight $\omega^a_{k|k}$. The maximum number of global hypothesis is $500$, and we prune global hypotheses with weight smaller than $10^{-3}$. We also prune Bernoulli components with existence probability smaller than $10^{-5}$ and Gaussian components in the Poisson intensity with weight smaller than $10^{-5}$. In addition, we consider $L$-scan implementation with $L=3$ \cite{garcia2019trajectory}, where control points before the last three time instants are considered independent and remain unchanged with new measurements. Furthermore, Bernoulli components with probability of being alive at the current time step smaller than $10^{-5}$ are considered dead, and they are no longer predicted or updated.

To report the traffic line estimates, we first find the TPMBM posterior density conditioned on the particle with the highest weight, and then we find the multi-Bernoulli component with the highest weight and its Bernoulli components with existence probability one. For each Bernoulli component, we extract the sequence of B-spline control points with the most likely length. Finally, we interpolate the control points to obtain continuous traffic line estimates. The vehicle pose estimate at each time step is obtained by taking the sample mean instead of choosing the particle with the highest weight at the last time step due to particle history degeneracy.

\begin{figure*}
    \centering
    \includegraphics[width=2\columnwidth]{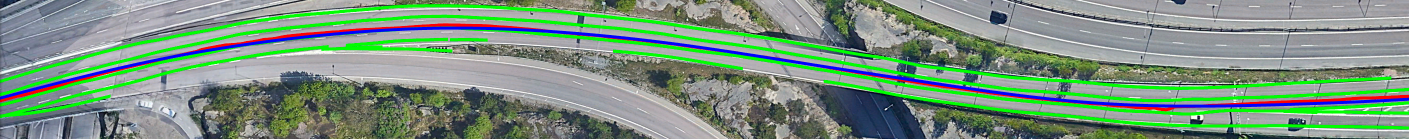}

    \vspace{0.03cm}
    \includegraphics[width=2\columnwidth]{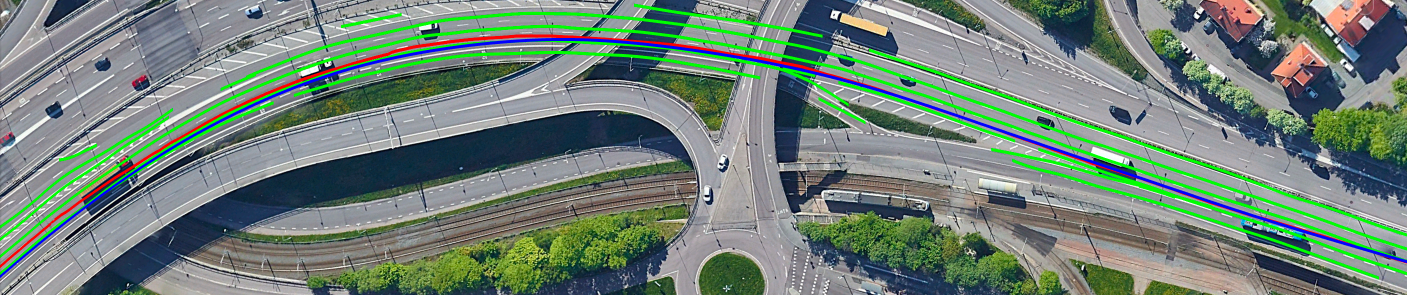}

    \vspace{0.03cm}
    \includegraphics[width=2\columnwidth]{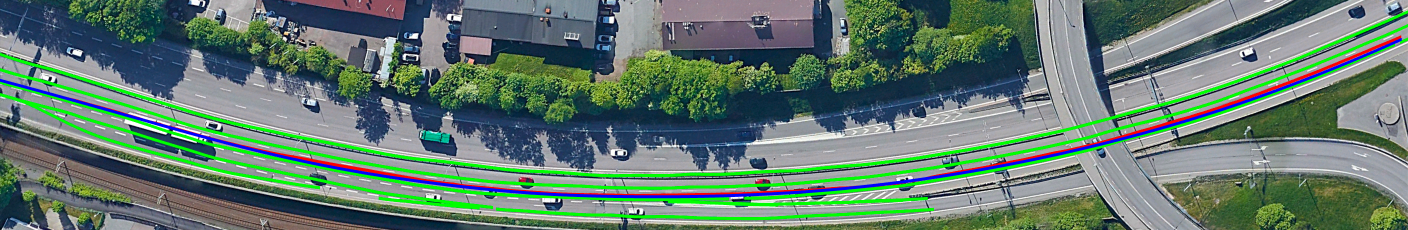}
    \caption{OxTS vehicle trajectory (\textcolor{blue}{blue}), estimated vehicle trajectory (\textcolor{red}{red}), estimated traffic lines (\textcolor{green}{green}) overlaid with Google Earth satellite image. The top figure illustrates the best traffic line estimation performance, followed by the bottom figure and the mid-figure. }
    \label{fig_multi_lane_estimation}
\end{figure*}

In the experiment, we use $N_p=1000$ particles, initialized only using information provided by the GNSS measurement. The multi-lane estimation performance is evaluated by overlaying the estimated traffic lines on the Google Earth satellite image; screenshots are provided in Fig. \ref{fig_multi_lane_estimation}. It can be observed from the results that the estimated traffic lines are continuous and smooth, and that they align well with the lane markings in the satellite image up to a lateral offset. This offset mainly depends on the lateral localization accuracy of the vehicle as in most cases, the offset between the estimated traffic lines and the lane markings matches the offset between the OxTS and the estimated vehicle trajectories. Lane splittings are also well estimated, with the only exception being at the bottom-left of Fig. \ref{fig_multi_lane_estimation}, where a newborn traffic line is estimated to be spawned from an existing one.

The root-mean-square localization error in the $x$, $y$, and $z$ directions are 0.97\,m, 1.10\,m, and 2.06\,m, respectively. The improvement with respect to using only visual odometry and GNSS is about 11\%. Since lane marking detections can only provide accurate relative lateral information and GNSS measurement is of low accuracy, it is difficult to accurately localize the vehicle and track the lanes without using detections that have good longitudinal information, such as traffic signs/lights. Nevertheless, our proposed solution can provide a reasonable estimate of traffic lines with respect to the vehicle trajectory.

% The root-mean-square-errors of vehicle pose estimates are presented in Table \ref{eq_performance_metrics}.
% \begin{table}[t]
%     \centering
%     \label{eq_performance_metrics}
%     \caption{The RMSE of vehicle pose estimates in meters and degrees.}
%     \begin{tabular}{c|cccccc}
%     \hline
%     & $x$ & $y$ & $z$ & yaw & pitch & roll \\ \hline
%     GNSS  &     &     &     &     &     &    \\
%     GNSS + Visual Odometry &     &     &     &     &     &    \\
%     SLAMOT  &     &     &     &     &     &     
%     \end{tabular}
% \end{table}

\section{Conclusion and Future Work}\label{sec_conclusion}

In this paper, we have proposed an effective solution for simultaneous localization and multi-lane tracking using onboard sensors and a SD map. The proposed solution is based on a TPMBM-based SLAMOT filter, where the estimation of traffic lines, in the form of B-spline trajectories, using lane marking detection points is formulated as a multiple extended object tracking problem. The efficacy of the proposed solution has been demonstrated on real-world data.

In the future, we plan to work on the batch formulation of the SLAMOT problem such that smoothed vehicle trajectory and traffic line estimates could be obtained by exploiting full information of the data. The mapping of static landmarks, such as traffic signs and traffic lights, can also be incorporated into the solution to further improve the localization accuracy, and hence also the multi-lane estimation performance.

\bibliographystyle{IEEEtran}
\bibliography{mybibli.bib}

\end{document}